\documentclass{article}
\usepackage{tap_arxiv,times}
\usepackage{amsmath,amssymb,booktabs}
\usepackage{algorithm,algpseudocode}
\usepackage{xcolor,colortbl}
\usepackage{graphicx,multirow,tabularx}
\usepackage{hyperref}
\hypersetup{hidelinks,
pdftitle={TAP: Efficient Long-Horizon Agent Pruning via Trajectory-Anchored Recovery},
pdfauthor={Yuanzhe Li, Pengxin Wang, Yuxin Ren, Jianing Deng, Jingtong Hu, Song Wang, Jingdi Chen, Huanrui Yang}}
\usepackage{url}

\title{TAP: Efficient Long-Horizon Agent Pruning via Trajectory-Anchored Recovery}
\author{\parbox{\dimexpr\textwidth-2\tabcolsep\relax}{\centering\normalfont
Yuanzhe Li\textsuperscript{1}, Pengxin Wang\textsuperscript{1}, Yuxin Ren\textsuperscript{1}, Jianing Deng\textsuperscript{2}\\[0.3em]
Jingtong Hu\textsuperscript{2}, Song Wang\textsuperscript{3}, Jingdi Chen\textsuperscript{1}, Huanrui Yang\textsuperscript{1}\\[0.7em]
{\small\textsuperscript{1}University of Arizona\\
\textsuperscript{2}University of Pittsburgh\\
\textsuperscript{3}University of Central Florida}
}}

\begin{document}
\maketitle
\lhead{}
\renewcommand{\headrulewidth}{0pt}
\begin{abstract}
Emerging long-horizon agentic tasks require repeated model calls, worsening the inference cost of already-costly language models. While narrow agentic tasks suggest potential for aggressive model pruning without performance drop, empirical results show existing methods proposed for question answering tasks severely degrade task performance when applied to agentic models. We trace this failure to two decisions: what to prune and how to recover. For pruning, one-shot importance estimates fail to track how the pruned model adapts. For recovery, offline distillation covers only teacher prefixes, while full-trajectory on-policy distillation causes student errors to compound across turns. In this work, we propose \emph{Trajectory-Anchored Pruning (TAP)}, the first structural pruning framework for reinforcement learning (RL)-trained agents.
TAP couples structural pruning with efficient on-policy recovery, anchoring interactions to teacher trajectories while allowing the student to generate each reasoning--action response. A frozen dense teacher supervises the student's response prefixes, addressing within-response training--inference mismatch while preventing student-induced deviations from propagating across training turns. Instead of one-shot pruning, TAP re-scores channels using gradients of the recovery objective on the recovered student, connecting iterative channel selection to the evolving policy. 
With 60\% of FFN channels removed, TAP retains 99.2\% and 88.0\% of the dense 7B agents' task success rates on ALFWorld and WebShop, respectively, while reducing GPU time per successful task by approximately 22\% and 17\%. These results demonstrate effective structural compression of long-horizon agents under a limited recovery budget.

\end{abstract}

\section{Introduction}
\label{sec:introduction}
\suppressfloats[t]

\begin{figure}[t]
\centering
\includegraphics[width=\linewidth]{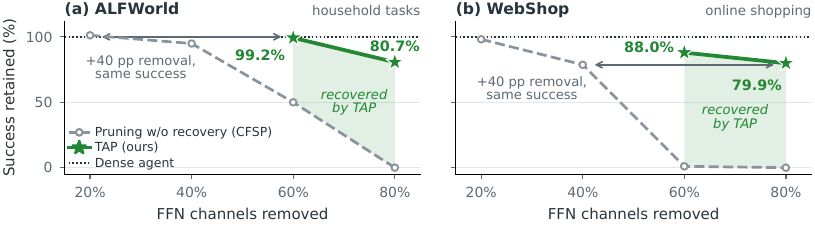}
\caption{\textbf{RL-trained agents are highly compressible, but only with suitable pruning and recovery.} Share of the dense agent's task success retained by 7B agents as FFN channels are removed. Pruning without recovery collapses beyond $40\%$ removal, whereas TAP retains about $80$--$99\%$ of dense success up to $80\%$ removal.}
\label{fig:pruning-sensitivity}
\end{figure}

Reinforcement learning (RL) has produced language agents with strong capabilities in multi-step tasks~\citep{feng2025gigpo,he2026hgpo,qi2026stapo,zhen2026stephrl,li2026soar,qin2026spear}. Each decision, however, conditions on environment observations, tool descriptions, and an ever-growing interaction history, and solving a long-horizon task requires many such model invocations, so inference cost accumulates over the trajectory~\citep{kang2026acon,wang2026swepruner,li2026dmr}. This growing cost motivates model compression for efficient agent deployment.

Compressing an LLM tuned for a specific agentic task should, in principle, be unusually forgiving. Compression needs to preserve only the decision-making the task requires, not the pretrained model's broad knowledge, which suggests room for more aggressive pruning than general-purpose LLMs tolerate~\citep{hu2025trimllm}.
However, existing structural pruning methods, developed and evaluated primarily on static tasks~\citep{ma2023llmpruner,zhang2024loraprune,muralidharan2024minitron,zhang2026shortopd}, fail to exploit it. Our experiments confirm that a 7B WebShop agent with 80\% of its FFN channels removed can retain the success rate attained at only 40\% removal by previous methods (Figure~\ref{fig:pruning-sensitivity}b), indicating a significant gap between the compression potential of agentic LLMs and what existing methods achieve.

We attribute this gap to limitations in two key decisions: what to prune and how to recover afterwards.
\textbf{Deciding what to prune.} Existing methods typically estimate channel importance once, on the dense model, and remove the selected channels in a single step. Two challenges arise when applying this recipe to RL-trained agents. First, what must be preserved is the agent's action-generating policy on interaction contexts, which criteria based on token likelihood only approximate; even on agent contexts, replacing a distribution-matching criterion with likelihood lowers out-of-distribution success by 5.5 points (Table~\ref{tab:calibration-loss}). Second, a single cut commits to a channel set before observing how the pruned student adapts. After recovery, re-scoring changes which channels are retained relative to one-shot selection (Table~\ref{tab:mask-recalibration}). In each case, pruning moves the student away from the dense agent in ways the selection criterion does not see, leading to teacher--student policy drift.

\textbf{Deciding how to recover.} Knowledge distillation is the standard way to recover structurally pruned models~\citep{muralidharan2024minitron}, but distillation on fixed teacher trajectories trains the student only on teacher prefixes, whereas at deployment the student conditions on its own outputs. In the agent setting this training--inference mismatch is severe: KD-recovered baselines steadily reduce their distillation loss, yet their steps produce missing or malformed action blocks. On-policy distillation (OPD) removes this mismatch by supervising the student's own generations~\citep{gu2024minillm,agarwal2024gkd}. For long-horizon agents, however, full-trajectory OPD also lets the student's actions determine subsequent environment states. A heavily pruned student then trains on states increasingly unlike those the teacher visits, where teacher supervision becomes less reliable~\citep{li2026rethinkingopd,wang2026tcod,li2026guidedopd}, and its errors compound across turns~\citep{wang2026tcod,li2026guidedopd}.

To address these two limitations, we propose \textbf{TAP}, an iterative pruning framework with \emph{trajectory-anchored on-policy distillation}. TAP fixes interaction contexts to teacher trajectories while allowing the student to generate each complete reasoning--action response under token-level supervision from a frozen dense teacher. This retains on-policy supervision within each turn while preventing student actions from altering subsequent training contexts. We interleave structural pruning with anchored recovery and, before each subsequent cut, re-score channels using gradients of the same recovery objective on the recovered student, coupling policy recovery with structural selection.

Experiments show that with 60\% of FFN channels removed, TAP retains 99.2\% and 88.0\% of the dense 7B agents' success rates on ALFWorld and WebShop, respectively, while reducing GPU time per successful task by approximately 22\% and 17\%; at 80\% removal, it retains 80.7\% and 79.9\%. TAP also runs efficiently, using only 0.18--1.21 million supervised response tokens across all recovery stages for the 1.5B and 7B agents.

Our main contributions are summarized as follows:
\begin{itemize}
    \item We study structural pruning of RL-trained long-horizon agents and trace existing methods' sharp degradation at aggressive ratios to limitations in \emph{what to prune} and \emph{how to recover}. We further show that decreasing distillation loss can coexist with malformed generation.
    \item We propose TAP, which couples trajectory-anchored OPD with iterative pruning and recovery-informed channel recalibration, so that the recovery objective also guides channel selection at each stage.
    \item Experiments on ALFWorld and WebShop at two model scales demonstrate improved task-success retention and reduced deployment cost, with ablations isolating the contributions of iterative pruning, recalibration, and anchoring.
\end{itemize}

\section{Related Work}
\label{sec:related}

\paragraph{Structural pruning for LLMs.}
Structural pruning removes channels, attention heads, or layers so that the pruned model runs with standard dense operations. Prior work scores structures with gradient- or activation-based importance and adaptive budget allocation \citep{ma2023llmpruner,zhang2024loraprune,wang2025cfsp,guo2025slimllm}, compensates for pruning errors with curvature-aware updates or local reconstruction \citep{vanderouderaa2024llmsurgeon,wei2024sobp,ling2024slimgpt,guo2025slimllm}, and recovers the pruned model with parameter-efficient fine-tuning or distillation \citep{ma2023llmpruner,zhang2024loraprune,muralidharan2024minitron,zhang2026shortopd}, mostly on static language tasks. Gradual pruning interleaves pruning with training \citep{zhu2017prune}, and for LLMs, multi-step pruning with retraining improves on one-shot removal at high compression \citep{zhang2024loraprune,pan2025adaptpruneradaptivestructuralpruning}. For agents, ACBench evaluates unstructured and 2:4 weight sparsity without recovery training \citep{dong2025acbench}, REAP prunes MoE experts for coding and tool use \citep{lasby2026reap}, and pruning has been combined with RL for embodied control: Before Parc Ferm\'e prunes driving LLMs during RL, and RLRC recovers compressed vision-language-action models with SFT followed by RL \citep{benfenati2026bpf,chen2026rlrc}. We instead study structural pruning of dense RL-trained agents that interact with their environments over long horizons. Whereas prior criteria score channels by language-modeling gradients or activations, TAP scores them with gradients of the teacher--student divergence that its recovery minimizes on agent contexts, and re-scores them on the recovered student before each subsequent cut to limit drift that compounds over long horizons.

\paragraph{Distillation and on-policy distillation.}
Distillation transfers capabilities, including those acquired through RL, from a teacher to a smaller student \citep{guo2025deepseekr1,yang2025qwen3technicalreport}. A student trained on fixed teacher demonstrations may reach sequences or states absent from them \citep{ross2011dagger}; MiniLLM and GKD address this by supervising the student's own generations \citep{gu2024minillm,agarwal2024gkd}. For multi-turn agents, TCOD and Guided-OPD report that student rollouts compound errors across turns and leave the teacher's reliable support, and mitigate this with a rollout-depth curriculum or decaying teacher intervention \citep{wang2026tcod,li2026guidedopd}; TurnOPD \citep{zhou2026turnopd} rebalances rollout and loss budgets across turns. ShortOPD recovers structurally pruned models with short-to-long rollouts on mathematics, code, and open-ended generation \citep{zhang2026shortopd}. These methods collect student rollouts in the environment or target non-agentic tasks. TAP recovers a pruned agent by anchoring every training context to teacher trajectories while supervising the student's own responses, requires no environment interaction, and couples this recovery with iterative channel selection. Anchoring keeps training contexts where teacher supervision is reliable and stops student errors from propagating across turns.

\section{Method}
\label{sec:method}

\begin{figure}[t]
    \centering
    \includegraphics[width=\linewidth]{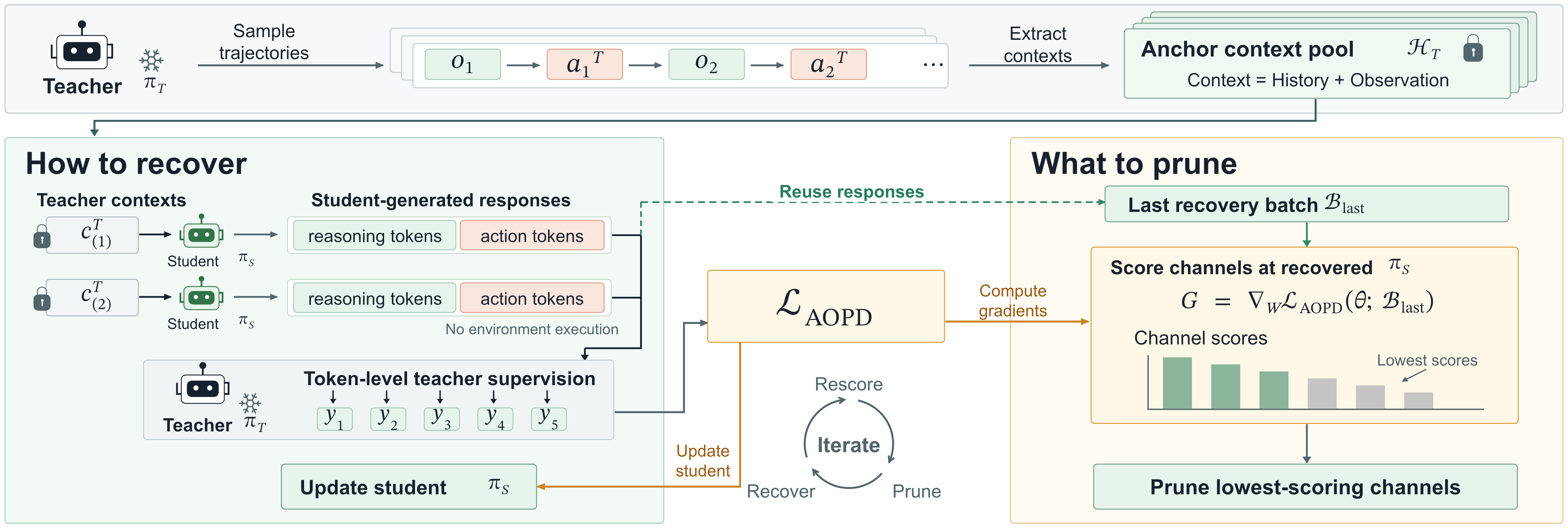}
    \caption{\textbf{TAP couples how to recover and what to prune through one objective.} Top: teacher trajectories provide a fixed pool of anchor contexts. Left (how to recover): within anchor contexts, the student generates reasoning and action tokens, the frozen teacher supervises every generated token, and minimizing $\mathcal L_{\mathrm{AOPD}}$ updates the student. Right (what to prune): gradients of the same objective on the last recovery batch score channels at the recovered student, and remove the lowest-scoring channels. The prune--recover--rescore cycle repeats until the target removal ratio is reached.}
    \label{fig:pruning-framework}
\end{figure}

\subsection{Preliminaries}
Let $x$ denote a task instruction and $o_t$ the observation at step $t$. Let $h_{<t}$ contain the task instruction and retained observation--action pairs before step $t$. The interaction context is $c_t=(h_{<t},o_t)$. We condition the policy on this context, writing $\pi_\theta(a_t\mid c_t)$ for brevity. This policy generates a response $a_t$ containing reasoning and an executable action. The environment executes the parsed action and returns $o_{t+1}$. We start from a dense agent $\pi_T$ after RL and freeze it as the teacher. We denote the student by $\pi_S=\pi_\theta$, where $\theta$ are its current parameters. Our objective is to preserve its complete-task performance under structural model reduction:
\begin{equation}
J(\pi_S)=\mathbb E_{x\sim\mathcal D_{\mathrm{eval}},\,\tau\sim\pi_S(\cdot\mid x)}[R(\tau)],
\label{eq:task-objective}
\end{equation}
where $\mathcal D_{\mathrm{eval}}$ is the evaluation task distribution, $\tau$ is a complete student--environment trajectory, and $R(\tau)$ is its task return.

We prune intermediate channels in SwiGLU FFNs by removing the corresponding rows of the gate and up projections and the corresponding column of the down projection. Formal definitions are provided in Appendix~\ref{app:ffn-removal}.

\paragraph{Overview.}
TAP progressively compresses the student over $K$ stages, coupling structural pruning with teacher-guided recovery (Figure~\ref{fig:pruning-framework}, Algorithm~\ref{alg:tap}). Each stage follows a prune--recover--rescore cycle: it removes FFN channels to reach a cumulative ratio $\rho_k$, recovers the pruned student by distillation from the frozen teacher within anchored contexts, and re-scores the remaining channels with gradients of the same recovery objective on the recovered student. This shared objective links the two steps: recovery restores the student's policy after each cut, and its objective, evaluated on its last batch of responses, scores channels for the next cut, while pruning targets the channels whose removal least perturbs the recovery objective. Together, they keep teacher supervision reliable by limiting drift from both pruning and long-horizon interaction. Section~\ref{sec:anchored-opd} introduces the recovery objective, and Section~\ref{sec:recovery-guided-pruning} explains how it guides iterative channel selection.

\begin{algorithm}[t]
\caption{TAP: Iterative Pruning with Trajectory-Anchored Recovery}
\label{alg:tap}
\small
\begin{algorithmic}[1]
\Require Teacher $\pi_T$, training tasks $\mathcal D_{\mathrm{train}}$, target ratio $\rho_\star$, stages $K$, token budget $Q$, warm-up fraction $\alpha$
\Statex \textbf{Notation:} $s_{\ell,j}$ is the importance score of FFN channel $j$ in layer $\ell$; $\mathcal B_{\mathrm{last}}$ is the last training batch of anchored OPD
\State Construct anchor pool $\mathcal H_T$ from teacher trajectories on $\mathcal D_{\mathrm{train}}$
\State Initialize $\pi_S\gets\pi_T$; keep teacher $\pi_T$ and anchor pool $\mathcal H_T$ fixed
\State $\{s_{\ell,j}\}\gets\operatorname{InitialScores}(\operatorname{GaussianPerturb}(\pi_S),\pi_T,\mathcal H_T)$
\For{$k=1,\ldots,K$}
    \State Set cumulative removal target $\rho_k$ using Eq.~\eqref{eq:cubic-schedule}
    \State \textbf{Prune:} $\pi_S\gets\operatorname{Prune}(\pi_S,\{s_{\ell,j}\},\rho_k)$
    \State \textbf{Recover:} $\pi_S\gets\operatorname{KDWarmup}(\pi_S,\pi_T,\mathcal H_T;\,\alpha Q/K)$
    \State \hphantom{\textbf{Recover:}} $(\pi_S,\mathcal B_{\mathrm{last}})\gets\operatorname{AnchoredOPD}(\pi_S,\pi_T,\mathcal H_T;\,(1-\alpha)Q/K)$
    \If{$k<K$}
        \State \textbf{Rescore:} compute OPD gradients on $\mathcal B_{\mathrm{last}}$ at the recovered $\pi_S$
        \State \hphantom{\textbf{Rescore:}} update $\{s_{\ell,j}\}$ using Eq.~\eqref{eq:channel-taylor-components} for the next cut
    \EndIf
\EndFor
\State \Return $\pi_S$
\end{algorithmic}
\end{algorithm}

\subsection{Trajectory-Anchored On-Policy Distillation}
\label{sec:anchors}
\label{sec:anchored-opd}

TAP recovers the pruned student with on-policy distillation anchored to teacher trajectories: the student generates its own responses, but only within interaction contexts taken from the teacher's trajectories.

\paragraph{Anchoring interaction context.}
We collect teacher trajectories on training tasks $\mathcal D_{\mathrm{train}}$ and extract a fixed pool $\mathcal H_T=\{(c_t^T,a_t^T)\}$ of context--response pairs. Each context $c_t^T=(h_{<t}^T,o_t^T)$ contains the retained teacher history and current observation; $a_t^T$ is the corresponding teacher response. We denote the sampling distribution over these contexts by $q_{\mathcal H_T}$. Given an anchor context formatted with the agent's prompt template, the current student generates a fresh response:
\begin{equation}
\widehat a_t\sim\pi_S(\cdot\mid c_t^T).
\label{eq:anchored-action}
\end{equation}
The generated action is not executed in the environment; the next training context is sampled from $\mathcal H_T$. This trajectory-anchored recovery requires no additional environment interaction during training.

\paragraph{Distillation on student responses.}
We construct each training batch $\mathcal B$ by sampling anchor contexts $c^T\sim q_{\mathcal H_T}$ and student responses $\widehat a$. For a response $\widehat a=(y_1,\ldots,y_L)$, let $p_j=(c^T,y_{<j})$ denote the context followed by the student-generated prefix. The frozen teacher scores the same prefixes, and we minimize the reverse KL averaged over the batch's $N_{\mathcal B}$ response tokens:
\begin{equation}
\mathcal L_{\mathrm{AOPD}}(\theta;\mathcal B)
=\frac{1}{N_{\mathcal B}}\sum_{(c^T,\widehat a)\in\mathcal B}\sum_{j=1}^{|\widehat a|}
D_{\mathrm{KL}}\!\left(\pi_\theta(\cdot\mid p_j)\,\|\,\pi_T(\cdot\mid p_j)\right).
\label{eq:aopd}
\end{equation}
Following GKD \citep{agarwal2024gkd}, sampled responses are held fixed during each update; following Guided-OPD \citep{li2026guidedopd}, we use reverse KL with supervision restricted to response tokens.

\paragraph{Relation to KD and full-trajectory OPD.}
Recovery objectives differ in where the supervised prefixes $p_j=(c,y_{<j})$ come from (definitions in Appendix~\ref{app:recovery-objectives}). Logits KD supervises teacher prefixes, pairing a teacher context with a teacher response prefix, so the student never learns to continue from its own tokens. Full-trajectory OPD supervises prefixes generated entirely by the student, so its contexts shift with the student's own actions. Anchored OPD takes the context from teacher trajectories and the response prefix from the student: supervision covers the prefixes the student produces within each turn, while its contexts stay where teacher supervision is reliable, and recovery requires no environment interaction.

\subsection{Recovery-Guided Iterative Pruning}
\label{sec:recovery-guided-pruning}
\label{sec:iterative-pruning}
\label{sec:opd-channel-recalibration}

\paragraph{Channel importance from the recovery objective.}
We select channels by minimal disturbance: a channel is unimportant if removing it barely changes the anchored OPD objective, the teacher--student divergence that recovery minimizes, so pruning aims to keep the student close to the teacher. Following loss-based pruning \citep{ma2023llmpruner}, we approximate this change with a first-order Taylor criterion. For layer $\ell$, let $W_\ell^g,W_\ell^u,W_\ell^d$ denote the student's gate, up, and down weights, and let $G_\ell^{p}=\partial\mathcal L_{\mathrm{AOPD}}(\theta;\mathcal B)/\partial W_\ell^p$ for $p\in\{g,u,d\}$ be the gradient of Eq.~\eqref{eq:aopd} on a calibration batch $\mathcal B$ of anchor contexts and student responses. The importance of intermediate channel $j$ combines the corresponding gate/up rows and down column:
\begin{equation}
s_{\ell,j}=\sum_i\left|W_{\ell,ji}^{g}G_{\ell,ji}^{g}\right|
+\sum_i\left|W_{\ell,ji}^{u}G_{\ell,ji}^{u}\right|
+\sum_i\left|W_{\ell,ij}^{d}G_{\ell,ij}^{d}\right|,
\label{eq:channel-taylor-components}
\end{equation}
where $i$ indexes the hidden dimension and weights use output-by-input indexing. We remove the lowest-scoring channels within each layer. Because the scores reuse recovery responses, they require no additional environment interaction.

At the first cut, the student equals the teacher and the KL gradients vanish. We therefore compute the initial scores on a perturbed student, obtained by adding small Gaussian noise to its FFN weights, and apply the resulting selection to the unperturbed student. Details of channel scoring and initial calibration are provided in Appendix~\ref{app:channel-score-aggregation}.

\paragraph{Iterative pruning.}
Pruning and recovery change the student's parameters and policy, so a ranking estimated once before compression no longer reflects the adapted student. We therefore prune in $K$ stages and, before each subsequent cut, re-score the remaining channels on the recovered student using the last batch of recovery responses. Recovery between cuts also restores teacher--student alignment before further structural reduction. Let $\rho_k$ be the cumulative fraction of original FFN channels removed after cut $k$, and $\rho_\star$ the final target. Following the cubic schedule of \citet{zhu2017prune}, we increase the cumulative removal ratio with larger cuts early and smaller cuts later:
\begin{equation}
\rho_k=\rho_\star\left[1-\left(1-\frac{k}{K}\right)^3\right],
\qquad k=1,\ldots,K.
\label{eq:cubic-schedule}
\end{equation}
We use $K=3$ in our experiments; for an $80\%$ final target, the cumulative removal ratios are approximately $56\%$, $77\%$, and $80\%$. The teacher and anchor pool remain fixed throughout. We divide the total supervised response-token budget $Q$ equally across stages; each stage allocates a fraction $\alpha$ of its budget to logits KD warm-up on teacher response prefixes, followed by anchored OPD. We set $\alpha=0.2$ in our experiments.

\section{Experiments}
\label{sec:experiments}
We organize our experiments around four research questions (RQs). \textbf{RQ1} (Section~\ref{sec:exp-main}): Does TAP preserve task success under aggressive structural pruning? RQ2 and RQ3 address the two decisions identified in Section~\ref{sec:introduction}. \textbf{RQ2} (Section~\ref{sec:exp-prune}, \emph{what to prune}): Does re-scoring channels with the recovery objective on the adapted student select better channels than a fixed ranking? \textbf{RQ3} (Section~\ref{sec:exp-recover}, \emph{how to recover}): How does trajectory anchoring compare with offline KD and full-trajectory OPD? Finally, we verify \textbf{RQ4} (Section~\ref{sec:exp-efficiency}): Does structural compression reduce the cost of completing agent tasks?

\subsection{Experimental Setup}
\label{sec:exp-setup}

\paragraph{Benchmarks and models.}
We conduct experiments on ALFWorld and WebShop, two representative benchmarks for long-horizon language agents \citep{shridhar2021alfworld,yao2022webshop}. We use RL-trained agents at two model scales, \textbf{1.5B} and \textbf{7B}, from SPEAR and GiGPO \citep{qin2026spear,feng2025gigpo}, with model and checkpoint details provided in Appendix~\ref{app:checkpoints}. The FFN channel-pruning ratios are \textbf{40\%} and \textbf{60\%} for the \textbf{1.5B} models, and \textbf{60\%} and \textbf{80\%} for the \textbf{7B} models.

\paragraph{Baselines.}
We compare TAP with LLM-Pruner \citep{ma2023llmpruner}, CFSP \citep{wang2025cfsp}, and SlimGPT \citep{ling2024slimgpt}, grouped by how they recover the pruned agent. LLM-Pruner and CFSP select channels once and recover with logits KD. We evaluate SlimGPT with its native local reconstruction and no additional recovery training to assess the effectiveness of reconstruction alone in our agent setting. We additionally continue GiGPO training of the 1.5B agents while removing one percentage point of FFN channels per optimizer step (online RL, denoted RL + Pruning; Appendix~\ref{app:joint-rl-pruning}). All methods use uniform layer-wise FFN pruning ratios. For offline recovery comparisons, we match the total supervised token budget and batch size. Other baseline adaptations and budget construction are provided in Appendix~\ref{app:implementation}.

\paragraph{Evaluation settings.}
We evaluate on both the in-distribution (In) and out-of-distribution (Out) splits of ALFWorld and the full 500-task test set of WebShop. We report task success rate over three seeds and performance retention; for ALFWorld, retention averages the two splits. The ablations in Sections~\ref{sec:exp-prune} and~\ref{sec:exp-recover} use the 7B agents.

\subsection{Main Results: Task Success under Aggressive Pruning}
\label{sec:exp-main}
\textbf{RQ1.} \textit{Does TAP preserve task success under aggressive structural pruning?}

\begin{table}[t]
\centering
\caption{\textbf{TAP retains the most task success, and its lead grows with the removal ratio.} We test whether TAP preserves the success of RL-trained agents under aggressive FFN pruning, compared with pipelines that recover by online RL (RL + Pruning), offline logits KD (+ KD), or local reconstruction alone (SlimGPT). Entries are success rates over three seeds and retention relative to the dense agent; ALFWorld retention averages the In and Out splits. \textbf{Bold}/\underline{underline}: best/second best per setting. At the milder ratio of each model, the strongest baselines stay within $3.5$ retention points of TAP; at the aggressive ratio, TAP's retention exceeds the best baseline's by $8$--$60$ points. The redundancy of task-specific agents is therefore exploitable, but only with suitable channel selection and recovery.}
\label{tab:main}
\renewcommand{\arraystretch}{0.92}

\small
\setlength{\tabcolsep}{3pt}

\begin{tabularx}{\linewidth}{
    c>{\raggedright\arraybackslash}Xccccc
}
\toprule
\multirow{2}{*}{$\rho$}
& \multirow{2}{*}{Method}
& \multicolumn{3}{c}{\textbf{ALFWorld}}
& \multicolumn{2}{c}{\textbf{WebShop}} \\
\cmidrule(lr){3-5}\cmidrule(lr){6-7}
&
& In SR (\%)
& Out SR (\%)
& Retention (\%)
& SR (\%)
& Retention (\%) \\
\midrule

\multicolumn{7}{l}{\textbf{Qwen2.5-1.5B}} \\
\midrule
$0\%$ & Dense RL
& $87.86_{\pm 2.86}$ & $83.83_{\pm 1.55}$ & $100$
& $77.93_{\pm 0.12}$ & $100$ \\
\cmidrule{1-7}

& RL + Pruning
& $36.19_{\pm 2.30}$ & $35.82_{\pm 2.69}$ & $41.96$
& $4.07_{\pm 1.22}$ & $5.22$ \\

& LLM-Pruner + KD
& $73.81_{\pm 4.65}$ & $61.94_{\pm 4.54}$ & $78.95$
& $14.00_{\pm 0.72}$ & $17.96$ \\

& CFSP + KD
& $\mathbf{82.62}_{\pm 3.67}$ & $\underline{75.12}_{\pm 0.86}$ & $\underline{91.82}$
& $39.73_{\pm 16.01}$ & $50.98$ \\

& SlimGPT (recon.)
& $0.00_{\pm 0.00}$ & $0.00_{\pm 0.00}$ & $0.00$
& $\underline{60.60}_{\pm 0.80}$ & $\underline{77.76}$ \\

\rowcolor{black!8}
\cellcolor{white}\multirow{-5}{*}{$40\%$}
& \textbf{TAP (ours)}
& $\underline{82.38}_{\pm 0.41}$ & $\mathbf{77.36}_{\pm 1.88}$ & $\mathbf{93.02}$
& $\mathbf{63.33}_{\pm 0.90}$ & $\mathbf{81.26}$ \\
\cmidrule{1-7}

& RL + Pruning
& $6.90_{\pm 2.30}$ & $1.74_{\pm 1.55}$ & $4.97$
& $0.00_{\pm 0.00}$ & $0.00$ \\

& LLM-Pruner + KD
& $30.00_{\pm 0.71}$ & $25.12_{\pm 1.55}$ & $32.06$
& $0.07_{\pm 0.12}$ & $0.09$ \\

& CFSP + KD
& $\underline{67.38}_{\pm 1.09}$ & $\underline{63.93}_{\pm 2.15}$ & $\underline{76.48}$
& $4.20_{\pm 2.60}$ & $5.39$ \\

& SlimGPT (recon.)
& $0.00_{\pm 0.00}$ & $0.00_{\pm 0.00}$ & $0.00$
& $\underline{7.07}_{\pm 1.33}$ & $\underline{9.07}$ \\

\rowcolor{black!8}
\cellcolor{white}\multirow{-5}{*}{$60\%$}
& \textbf{TAP (ours)}
& $\mathbf{78.81}_{\pm 4.59}$ & $\mathbf{67.66}_{\pm 2.83}$ & $\mathbf{85.21}$
& $\mathbf{54.07}_{\pm 1.53}$ & $\mathbf{69.38}$ \\

\midrule

\multicolumn{7}{l}{\textbf{Qwen2.5-7B}} \\
\midrule
$0\%$ & Dense RL
& $94.29_{\pm 1.89}$ & $89.55_{\pm 0.75}$ & $100$
& $78.07_{\pm 0.50}$ & $100$ \\
\cmidrule{1-7}

& LLM-Pruner + KD
& $84.05_{\pm 0.41}$ & $75.87_{\pm 1.72}$ & $86.92$
& $52.47_{\pm 0.50}$ & $67.21$ \\

& CFSP + KD
& $\mathbf{92.62}_{\pm 1.49}$ & $\underline{84.58}_{\pm 1.88}$ & $\underline{96.34}$
& $\underline{66.27}_{\pm 0.76}$ & $\underline{84.88}$ \\

& SlimGPT (recon.)
& $\underline{86.43}_{\pm 1.24}$ & $83.08_{\pm 3.53}$ & $92.22$
& $51.60_{\pm 1.22}$ & $66.10$ \\

\rowcolor{black!8}
\cellcolor{white}\multirow{-4}{*}{$60\%$}
& \textbf{TAP (ours)}
& $\mathbf{92.62}_{\pm 1.65}$ & $\mathbf{89.80}_{\pm 1.55}$ & $\mathbf{99.24}$
& $\mathbf{68.67}_{\pm 0.95}$ & $\mathbf{87.96}$ \\
\cmidrule{1-7}

& LLM-Pruner + KD
& $26.43_{\pm 1.89}$ & $21.64_{\pm 1.49}$ & $26.06$
& $25.60_{\pm 0.69}$ & $32.79$ \\

& CFSP + KD
& $\underline{66.19}_{\pm 2.18}$ & $\underline{63.18}_{\pm 3.11}$ & $\underline{70.38}$
& $\underline{56.00}_{\pm 1.40}$ & $\underline{71.73}$ \\

& SlimGPT (recon.)
& $6.43_{\pm 3.78}$ & $2.24_{\pm 1.49}$ & $4.62$
& $0.00_{\pm 0.00}$ & $0.00$ \\

\rowcolor{black!8}
\cellcolor{white}\multirow{-4}{*}{$80\%$}
& \textbf{TAP (ours)}
& $\mathbf{76.19}_{\pm 1.80}$ & $\mathbf{72.14}_{\pm 3.11}$ & $\mathbf{80.66}$
& $\mathbf{62.33}_{\pm 1.21}$ & $\mathbf{79.85}$ \\
\bottomrule
\end{tabularx}
\end{table}

\paragraph{Comparison with pruning baselines.}
Table~\ref{tab:main} shows that the baselines remain competitive only at the milder ratio of each model: at $60\%$ removal of the 7B agents, CFSP + KD retains $96.34\%$ of dense ALFWorld success, versus $99.24\%$ for TAP. The gap opens at the aggressive ratio. At $60\%$ removal of the 1.5B agents, no baseline retains more than $9.07\%$ of dense WebShop success, whereas TAP retains $69.38\%$; at $80\%$ removal of the 7B agents, TAP exceeds CFSP + KD by $10.28$ and $8.12$ retention points on ALFWorld and WebShop. The other recovery routes are brittle. Pruning during RL retains at most $41.96\%$ of dense success already at $40\%$ removal. SlimGPT's reconstruction alone collapses on 1.5B ALFWorld and on the 7B agents at $80\%$, despite near-dense success at $20\%$ (Appendix~\ref{app:slimgpt-15b}).

\paragraph{Pruning sensitivity.}
Figure~\ref{fig:pruning-sensitivity} (Section~\ref{sec:introduction}) shows how much of this capacity is recoverable. Without recovery, CFSP success declines sharply with channel removal: WebShop SR falls to $0.8\%$ at $60\%$ removal, and both benchmarks reach zero at $80\%$. TAP retains a similar share of dense success at substantially higher removal: $99.2\%$ on ALFWorld at $60\%$ removal, comparable to prune-only CFSP at $20\%$, and $79.9\%$ on WebShop at $80\%$, comparable to prune-only CFSP at $40\%$.

\subsection{What to Prune: Recovery-Guided Channel Recalibration}
\label{sec:exp-prune}
\textbf{RQ2.} \textit{Does re-scoring channels with the recovery objective on the adapted student select better channels than a fixed ranking?}

\begin{table}[t]
\centering
\caption{\textbf{Iterative pruning helps when channels are re-scored with the recovery objective.} To isolate what to prune as the student adapts, we fix the recovery objective and vary how channels are removed on Qwen2.5-7B. Cubic and uniform schedules use three pruning stages. A and B delete the same final channels, so B vs.\ A isolates interleaving recovery between cuts; B vs.\ C isolates recalibration; U vs.\ C compares uniform and cubic schedules. \textbf{Bold}/\underline{underline}: best/second best. Iterating with a fixed ranking gives mixed changes, whereas recalibration improves five of six metrics, and C improves on one-shot pruning in all six. The gain therefore comes from coupling channel selection with recovery.}
\label{tab:stages}
\renewcommand{\arraystretch}{0.92}
\footnotesize
\setlength{\tabcolsep}{1.8pt}
\begin{tabular*}{\linewidth}{@{\extracolsep{\fill}}lcccccccc@{}}
\toprule
& & & \multicolumn{3}{c}{\textbf{$60\%$ FFN removal}} & \multicolumn{3}{c}{\textbf{$80\%$ FFN removal}} \\
\cmidrule(lr){4-6}\cmidrule(lr){7-9}
Variant & Schedule & Ranking & ALF In & ALF Out & WebShop & ALF In & ALF Out & WebShop \\
\midrule
A & One-shot & Fixed & $\underline{92.38}_{\pm 2.51}$ & $86.32_{\pm 2.40}$ & $\underline{66.87}_{\pm 0.23}$ & $73.81_{\pm 2.06}$ & $61.69_{\pm 2.28}$ & $57.40_{\pm 1.22}$ \\
B & Cubic & Fixed & $90.48_{\pm 1.65}$ & $87.81_{\pm 0.86}$ & $63.27_{\pm 0.12}$ & $\mathbf{77.38}_{\pm 2.70}$ & $\underline{63.93}_{\pm 1.14}$ & $\underline{57.93}_{\pm 1.01}$ \\
\rowcolor{black!8}
\textbf{C (TAP)} & Cubic & Recalibrated & $\mathbf{92.62}_{\pm 1.65}$ & $\mathbf{89.80}_{\pm 1.55}$ & $\mathbf{68.67}_{\pm 0.95}$ & $\underline{76.19}_{\pm 1.80}$ & $\mathbf{72.14}_{\pm 3.11}$ & $\mathbf{62.33}_{\pm 1.21}$ \\
\midrule
U & Uniform & Recalibrated & $90.71_{\pm 1.24}$ & $\underline{89.55}_{\pm 0.75}$ & $65.80_{\pm 1.11}$ & $50.00_{\pm 1.89}$ & $43.03_{\pm 0.86}$ & $51.93_{\pm 1.92}$ \\
\bottomrule
\end{tabular*}
\end{table}

\paragraph{Iterative pruning and channel recalibration.}
In Table~\ref{tab:stages} (protocol in Appendix~\ref{app:iterative-ablation-protocol}), iterative pruning with a fixed ranking (B) gives mixed changes relative to one-shot pruning (A), lowering two of three metrics at $60\%$ removal. Recalibration makes iteration pay off: C improves on B in five of six metrics, by up to $8.21$ points on ALFWorld Out at $80\%$ removal, and improves on one-shot pruning in all six. Recalibration also changes the surviving subnetwork, replacing $14$--$22\%$ of the channels that one-shot selection retains (Appendix~\ref{app:mask-recalibration}). With recalibration, a uniform schedule (U) trails the cubic one on all six metrics.

\paragraph{The objective used for recalibration.}
Replacing TAP's reverse-KL scoring (Section~\ref{sec:opd-channel-recalibration}) with token NLL leaves ALFWorld In SR similar but lowers Out SR from $89.80\%$ to $84.33\%$ at $60\%$ removal (Table~\ref{tab:calibration-loss}), so scoring channels by the teacher--student divergence better preserves the agent's policy. Overall, iterative pruning pays off consistently only when channel importance is re-estimated with the recovery objective on the adapted student.

\subsection{How to Recover: Trajectory-Anchored Distillation}
\label{sec:exp-recover}
\textbf{RQ3.} \textit{How does trajectory anchoring compare with offline KD and full-trajectory OPD?}

\begin{table}[h]
\centering
\caption{\textbf{Recovery works best when the student generates its own responses within teacher interaction contexts.} To isolate how a pruned agent should be recovered, we fix the pruning path to TAP's and vary only the recovery objective on Qwen2.5-7B. \emph{Context} and \emph{Resp.}\ (response) indicate whose interaction contexts the student conditions on and whose responses it is supervised on: logits KD and SeqKD use teacher responses with soft and hard targets, full-trajectory OPD uses the student's own rollouts, and anchored OPD uses student responses within teacher contexts. Entries are SR; \textbf{bold}/\underline{underline}: best/second best. Anchored OPD is the best or within one standard deviation of the best in every column, whereas full-trajectory OPD collapses on ALFWorld at $80\%$ removal. Comparing rows isolates the two design choices: supervising the student's own responses improves on KD, and anchoring its contexts avoids the collapse of full-trajectory OPD.}
\label{tab:recovery}
\renewcommand{\arraystretch}{0.92}
\footnotesize
\setlength{\tabcolsep}{2pt}
\begin{tabular*}{\linewidth}{@{\extracolsep{\fill}}lcccccccc@{}}
\toprule
& \multicolumn{2}{c}{Source} & \multicolumn{3}{c}{\textbf{$60\%$ FFN removal}} & \multicolumn{3}{c}{\textbf{$80\%$ FFN removal}} \\
\cmidrule(lr){2-3}\cmidrule(lr){4-6}\cmidrule(lr){7-9}
Recovery & Context & Resp. & ALF In & ALF Out & WebShop & ALF In & ALF Out & WebShop \\
\midrule
D: Logits KD & Teacher & Teacher & $\mathbf{93.33}_{\pm 1.65}$ & $\underline{88.06}_{\pm 1.29}$ & $\underline{65.60}_{\pm 0.72}$ & $\underline{75.95}_{\pm 2.18}$ & $\underline{70.90}_{\pm 5.38}$ & $59.80_{\pm 0.80}$ \\
F: SeqKD & Teacher & Teacher & $\underline{92.62}_{\pm 1.09}$ & $85.32_{\pm 0.86}$ & $61.87_{\pm 0.58}$ & $70.95_{\pm 4.86}$ & $65.42_{\pm 3.76}$ & $55.40_{\pm 1.60}$ \\
E: Traj.\ OPD & Student & Student & $92.14_{\pm 0.71}$ & $86.07_{\pm 1.55}$ & $64.07_{\pm 0.23}$ & $49.76_{\pm 2.89}$ & $41.54_{\pm 5.80}$ & $\mathbf{62.67}_{\pm 0.61}$ \\
\rowcolor{black!8}
\textbf{C: Anchored} & Teacher & Student & $\underline{92.62}_{\pm 1.65}$ & $\mathbf{89.80}_{\pm 1.55}$ & $\mathbf{68.67}_{\pm 0.95}$ & $\mathbf{76.19}_{\pm 1.80}$ & $\mathbf{72.14}_{\pm 3.11}$ & $\underline{62.33}_{\pm 1.21}$ \\
\bottomrule
\end{tabular*}
\end{table}

\paragraph{Offline versus on-policy supervision.}
In Table~\ref{tab:recovery}, with teacher contexts fixed, supervising the student's own responses (anchored OPD) rather than teacher responses improves WebShop SR by $3.07$ and $2.53$ points over logits KD and by $6.80$ and $6.93$ points over SeqKD at $60\%$ and $80\%$ removal, and gives the highest ALFWorld Out SR at both ratios, while logits KD remains comparable on ALFWorld In, where the training--inference mismatch~\citep{gu2024minillm,agarwal2024gkd} is likely smaller. Offline KD can fail more severely after aggressive one-shot pruning: on 1.5B WebShop at $60\%$ removal, the one-shot KD baselines reduce their KL by about $84\%$ while $87$--$93\%$ of their steps contain malformed action blocks (Appendix~\ref{app:spear-kd-failure}). In Table 3, all recovery variants use the same pruning masks as TAP and differ only in the recovery objective. With these masks, logits KD is already much stronger than one-shot CFSP + KD; TAP's gains over the baselines thus combine a better pruning path with anchored recovery.

\paragraph{Anchored versus full-trajectory OPD.}
Full-trajectory OPD also supervises the student's own responses, but lets its executed actions determine the interaction contexts. It trails anchored OPD on all three metrics at $60\%$ removal and collapses on ALFWorld at $80\%$ ($49.76\%$ and $41.54\%$ In and Out SR, versus $76.19\%$ and $72.14\%$). These failures are not malformed outputs: only $1.32\%$ of its steps contain a broken action block, and $95\%$ of failed episodes run to the maximum step limit, so the student acts validly but fails to finish the task. The collapse arises for the weakest student on ALFWorld's much longer episodes, consistent with the cross-turn drift reported for multi-turn OPD~\citep{wang2026tcod,li2026guidedopd}: the student's own rollouts steer supervision toward states unlike those in teacher trajectories, and errors compound over more turns. Anchoring interaction contexts to teacher trajectories thus keeps on-policy recovery stable for heavily pruned students on long horizons.

\subsection{Deployment Efficiency}
\label{sec:exp-efficiency}
\textbf{RQ4.} \textit{Does structural compression reduce the cost of completing agent tasks?}

\begin{table}[h]
\centering
\caption{\textbf{Compression lowers the cost per successful task when success is retained.} Agent cost depends on speed, steps per task, and success, so we report all three for dense and TAP-pruned Qwen2.5-7B agents. Steps: agent steps per task; E2E: end-to-end speedup per task; CPS: GPU time per success, including failed attempts, relative to dense (Appendix~\ref{app:efficiency}). CPS falls by $17$--$29\%$ except on ALFWorld at $80\%$ removal, where $1.8\times$ more steps per task outweigh faster inference.}
\label{tab:efficiency}
\renewcommand{\arraystretch}{1.2}
\small
\setlength{\tabcolsep}{4.5pt}
\begin{tabular}{lccccccccc}
\toprule
& & \multicolumn{2}{c}{\textbf{Speedup}} & \multicolumn{3}{c}{\textbf{ALFWorld}} & \multicolumn{3}{c}{\textbf{WebShop}} \\
\cmidrule(lr){3-4}\cmidrule(lr){5-7}\cmidrule(lr){8-10}
Model & Params & Prefill $\uparrow$ & Decode $\uparrow$ & Steps $\downarrow$ & E2E $\uparrow$ & CPS $\downarrow$ & Steps $\downarrow$ & E2E $\uparrow$ & CPS $\downarrow$ \\
\midrule
Dense & 7.62B & $1.00\times$ & $1.00\times$ & $11.8$ & $1.00\times$ & $1.00\times$ & $6.2$ & $1.00\times$ & $1.00\times$ \\
\rowcolor{black!8}
TAP, $\rho{=}60\%$ & 4.19B & $1.71\times$ & $1.73\times$ & $13.6$ & $\mathbf{1.30\times}$ & $\mathbf{0.78\times}$ & $7.0$ & $1.38\times$ & $0.83\times$ \\
\rowcolor{black!8}
TAP, $\rho{=}80\%$ & 3.05B & $\mathbf{1.94\times}$ & $\mathbf{2.26\times}$ & $21.6$ & $0.77\times$ & $1.61\times$ & $6.1$ & $\mathbf{1.76\times}$ & $\mathbf{0.71\times}$ \\
\bottomrule
\end{tabular}
\end{table}

\paragraph{Model inference efficiency.}
Structural removal translates into practical acceleration: in Table~\ref{tab:efficiency}, $60\%$ and $80\%$ FFN removal reduce the parameter count from $7.62$B to $4.19$B and $3.05$B, and yield $1.71\times$ and $1.94\times$ prefill speedups and $1.73\times$ and $2.26\times$ decoding speedups, respectively. The pruned models also need $45\%$ and $60\%$ less memory for their weights.

\paragraph{Task-level efficiency.}
These gains extend to complete agent evaluations: at $60\%$ removal, end-to-end speedups reach $1.30\times$ on ALFWorld and $1.38\times$ on WebShop; at $80\%$ removal, WebShop achieves a $1.76\times$ speedup. Thus, structural compression accelerates agent execution even with the additional costs of environment interaction. Because an agent calls the model at every step, per-call speedups accumulate over each trajectory. \textbf{Cost per success} (CPS), which also charges failed attempts, falls at $60\%$ removal to $0.78\times$ the dense baseline on ALFWorld and $0.83\times$ on WebShop, reducing the resource cost per success by approximately $22\%$ and $17\%$. WebShop at $80\%$ removal achieves the largest saving, lowering CPS to $0.71\times$, or approximately $29\%$ less GPU time per successful task. The exception is ALFWorld at $80\%$ removal, where $1.8\times$ more steps per task outweigh the faster inference.

\section{Conclusion}
\label{sec:conclusion}
We studied structural pruning of RL-trained long-horizon agents and traced the failure of existing pipelines at aggressive ratios to two decisions: what to prune and how to recover. TAP couples the two through a shared objective: it recovers each pruned student with trajectory-anchored on-policy distillation, supervising the student's own responses within teacher interaction contexts, and re-scores channels with gradients of this objective before each cut. With $60\%$ of FFN channels removed, TAP retains $99.2\%$ and $88.0\%$ of the dense 7B agents' success on ALFWorld and WebShop while lowering cost per successful task; with $80\%$ removed, it still retains $80.7\%$ and $79.9\%$.

Our analyses separate the two decisions. For recovery, KD loss can decrease while generated actions remain malformed, and full-trajectory OPD collapses for heavily pruned students on the longer-horizon environment, whereas anchored recovery avoids both. For channel selection, iterative pruning helps consistently only when channels are re-scored on the recovered student.

TAP relies on informative teacher trajectories, and anchored recovery does not cover every state the deployed student may visit. We study FFN channel pruning of Qwen2.5 agents on two environments; extending TAP to other architectures, pruning granularities, and agentic domains, together with adaptive anchor selection and a gradual transition to unanchored recovery, are natural next steps.

\section*{Acknowledgments}
This work used the Delta and DeltaAI systems at the National Center for Supercomputing
Applications (NCSA) [awards OAC 2005572 and OAC 2320345] through allocation CIS260333 from the
Advanced Cyberinfrastructure Coordination Ecosystem: Services \& Support (ACCESS)
program~\citep{boerner2023access}, which is supported by U.S. National Science
Foundation grants \#2138259, \#2138286, \#2138307, \#2137603, and \#2138296.

\clearpage
\appendix
\section{Pruning with Continued RL}
\label{app:joint-rl-pruning}
\begin{figure}[t]
\centering
\includegraphics[width=0.5\linewidth]{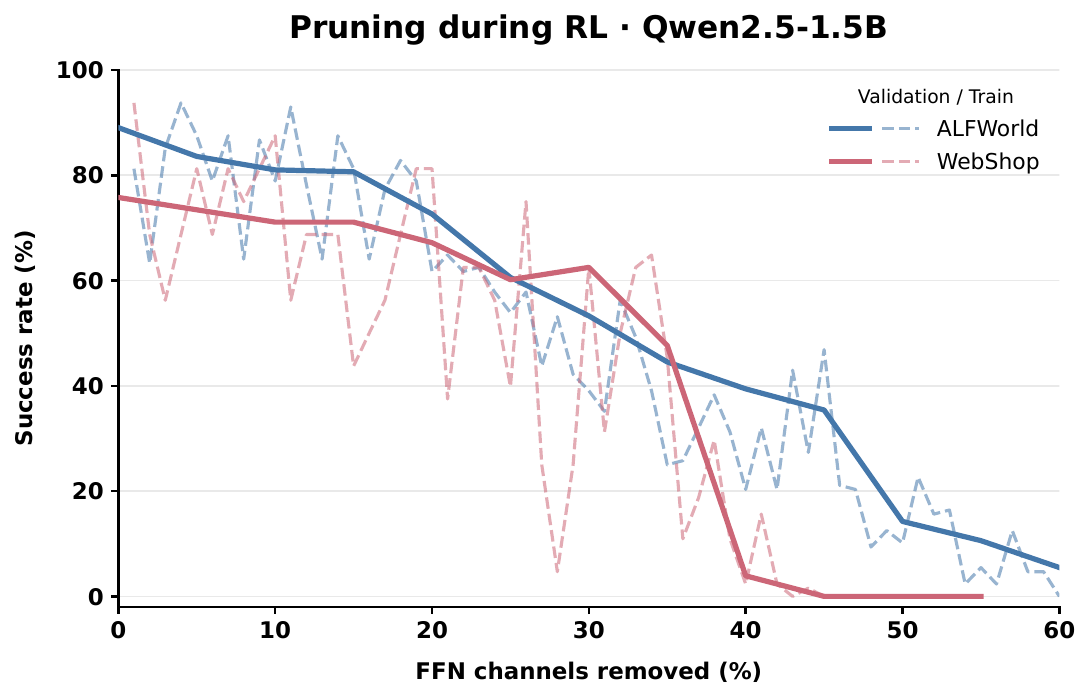}
\caption{Pruning during GiGPO continuation on Qwen2.5-1.5B agents, which removes one additional percentage point of FFN channels per optimizer step. Solid lines show periodic validation and dashed lines training-batch success. ALFWorld validation pools the 140 In and 134 Out tasks; WebShop uses 128 fixed tasks. Validation uses temperature $0.4$. Validation success falls as removal increases and drops sharply beyond about $30\%$ removal on WebShop, so continued RL does not keep the pruned agent functional under this schedule.}
\label{fig:rl-pruning}
\end{figure}

We evaluate pruning during RL on the SPEAR Qwen2.5-1.5B checkpoints using GiGPO continuation with rank-8 LoRA and learning rate $10^{-5}$. Starting from the dense checkpoint, each optimizer step removes an additional one percentage point of FFN channels using the baseline CFSP channel score. Channel masks are enforced during training. Online RL interacts with the environment and is not bound by the distillation-token budget. Figure~\ref{fig:rl-pruning} shows success during training, and Table~\ref{tab:main} reports the evaluations at $40\%$ and $60\%$ removal.

\section{Supplementary Results}
\subsection{SlimGPT at Lower Pruning Ratios on SPEAR-1.5B}
\label{app:slimgpt-15b}
Table~\ref{tab:slimgpt-15b-r02} reports the $20\%$ FFN-removal result on the SPEAR ALFWorld 1.5B checkpoint, using SlimGPT local reconstruction. At $40\%$ removal, the FFN width decreases from 8960 to 5376, and In and Out success rates are zero; these results appear in Table~\ref{tab:main}.
\begin{table}[t]
\centering
\caption{SlimGPT on SPEAR ALFWorld 1.5B at $20\%$ FFN channel removal. Success rates are percentages.}
\label{tab:slimgpt-15b-r02}
\small
\begin{tabular}{cccc}
\toprule
$\rho$ & FFN width & Split & Mean $\pm$ std \\
\midrule
$20\%$ & $8960\rightarrow7168$ & In & $87.62\pm2.51$ \\
$20\%$ & $8960\rightarrow7168$ & Out & $79.85\pm2.59$ \\
\bottomrule
\end{tabular}
\end{table}

\subsection{Channel-Mask Changes under Recalibration}
\label{app:mask-recalibration}
We compare the final channel masks from three-stage cubic pruning with recalibration (C) against one-shot selection from the same initial perturbed-student Taylor ranking (A), at matched final removal ratios, using the Qwen2.5-7B runs of Table~\ref{tab:stages}.

Let $U$ be the set of all eligible FFN channels, indexed by layer and original dense-model channel index, and let $D_A,D_C\subset U$ denote the equally sized deleted-channel sets. We report their Jaccard similarity $J_D=|D_A\cap D_C|/|D_A\cup D_C|$, the deleted-channel replacement fraction $R_D=|D_A\setminus D_C|/|D_A|$, and the retained-channel replacement fraction $R_K=|K_A\setminus K_C|/|K_A|$, where $K_A=U\setminus D_A$ and $K_C=U\setminus D_C$. Counts are pooled across layers before taking ratios.

\begin{table}[t]
\centering
\caption{Final channel masks of TAP (C) versus one-shot selection from the same initial ranking (A) on Qwen2.5-7B. $J_D$ is the Jaccard similarity of the deleted sets; $R_D$ and $R_K$ are the fractions of A's deleted and retained channels that C replaces. All values are percentages. Although the deleted sets largely overlap, recalibration replaces $14$--$22\%$ of the retained channels.}
\label{tab:mask-recalibration}
\small
\begin{tabular}{lcccc}
\toprule
Environment & Removal $\rho$ & Jaccard $J_D$ & Deleted $R_D$ & Retained $R_K$ \\
\midrule
ALFWorld & $60\%$ & $82.66$ & $9.49$ & $14.24$ \\
WebShop  & $60\%$ & $82.03$ & $9.87$ & $14.81$ \\
ALFWorld & $80\%$ & $89.44$ & $5.58$ & $22.31$ \\
WebShop  & $80\%$ & $90.13$ & $5.19$ & $20.76$ \\
\bottomrule
\end{tabular}
\end{table}

\begin{table}[t]
\centering
\caption{Effect of the calibration objective on Qwen2.5-7B at $60\%$ FFN channel removal on ALFWorld. NLL replaces KL for channel-importance estimation while the rest of TAP is unchanged. KL yields similar in-distribution but higher out-of-distribution success.}
\label{tab:calibration-loss}
\small
\begin{tabular}{lcc}
\toprule
Calibration objective & In SR (\%) & Out SR (\%) \\
\midrule
KL (TAP) & $\mathbf{92.62}_{\pm 1.65}$ & $\mathbf{89.80}_{\pm 1.55}$ \\
NLL & $91.90_{\pm 1.80}$ & $84.33_{\pm 2.99}$ \\
\bottomrule
\end{tabular}
\end{table}

\subsection{Decreasing KD Loss Does Not Ensure Executable Agent Behavior}
\label{app:spear-kd-failure}
We inspect the WebShop Qwen2.5-1.5B agents at $60\%$ FFN removal, where the one-shot KD baselines and TAP share a budget of 0.35M supervised response tokens (Table~\ref{tab:token-budgets}). For CFSP+KD, the response-token-weighted forward KL over the first and last 20 updates decreases from 1.580 to 0.250; for LLM-Pruner+KD, it decreases from 2.062 to 0.326.

\begin{table}[t]
\centering
\caption{Free-generation failure diagnostics for the same WebShop 1.5B setting. Percentages are computed on the 500 WebShop test goals. Action-block and invalid-action rates are computed over steps; repeated-action and step-limit rates are computed over episodes. Repeated action denotes at least three consecutive identical executed actions within an episode. Flags can overlap.}
\label{tab:spear-kd-failure}
\small
\begin{tabularx}{\linewidth}{Xrrrr}
\toprule
 & \multicolumn{2}{c}{Steps (\%)} & \multicolumn{2}{c}{Episodes (\%)} \\
Method & \shortstack{Missing/broken\\action block} & \shortstack{Invalid\\action} & \shortstack{Repeated\\action} & \shortstack{15-step\\limit} \\
\midrule
CFSP+KD & 87 & 95 & 58 & 84 \\
LLM-Pruner+KD & 93 & 93 & 82 & 98 \\
TAP & $<1$ & $<1$ & 4 & 4 \\
\bottomrule
\end{tabularx}
\end{table}

Despite decreasing teacher-forced loss, the baselines frequently fail to produce complete, non-empty \texttt{<action>} blocks, and the environment marks most executed actions invalid (Table~\ref{tab:spear-kd-failure}). These observations show that improved fitting under teacher prefixes can coexist with malformed free generation.

\subsection{Iterative Pruning Ablation Protocol}
\label{app:iterative-ablation-protocol}
In Table~\ref{tab:stages}, all variants share the dense initialization, initial response-only reverse-KL Taylor scores, anchor-context pool, final removal ratio, and total recovery-token budget $Q$. A uses $20\%$ of $Q$ for KD and $80\%$ for anchored OPD after its single cut. B, C, and U divide $Q$ equally across three iterations. Each iteration uses $20\%$ KD warm-up followed by $80\%$ anchored OPD. U uses cumulative removal ratios $\rho/3,2\rho/3,\rho$.

\section{Implementation Details}
\label{app:implementation}

\subsection{Dense Checkpoints and Evaluation Protocol}
\label{app:checkpoints}
We use the following author-released, environment-specific checkpoints:
\begin{itemize}
\item Qwen2.5-1.5B-Instruct (ALFWorld): \nolinkurl{yolay/SPEAR-ALFWorld-DrBoT-GiGPO-1.5B}.
\item Qwen2.5-1.5B-Instruct (WebShop): \nolinkurl{yolay/SPEAR-WebShop-DrBoT-GiGPO-1.5B}.
\item Qwen2.5-7B-Instruct (ALFWorld): \nolinkurl{langfeng01/GiGPO-Qwen2.5-7B-Instruct-ALFWorld}.
\item Qwen2.5-7B-Instruct (WebShop): \nolinkurl{langfeng01/GiGPO-Qwen2.5-7B-Instruct-WebShop}.
\end{itemize}
The 1.5B checkpoints use SPEAR + Dr.BoT (GiGPO), incorporating self-imitation and trajectory replay, whereas 7B uses GiGPO \citep{qin2026spear,feng2025gigpo}; cross-size differences also reflect different RL recipes. Evaluation episodes are capped at 50 steps on ALFWorld and 15 steps on WebShop, and decoding uses temperature $0.4$.

\subsection{FFN Channel Removal and Pruning Ratio}
\label{app:ffn-removal}

For a gated FFN at layer $\ell$, write
\begin{equation}
u_\ell(z)=\sigma(W_\ell^gz)\odot(W_\ell^uz),
\qquad \operatorname{FFN}_\ell(z)=W_\ell^du_\ell(z),
\end{equation}
where $\sigma$ is the SiLU activation, $W_\ell^g,W_\ell^u\in\mathbb R^{m_\ell\times d}$ and $W_\ell^d\in\mathbb R^{d\times m_\ell}$. Removing intermediate channel $j$ deletes row $j$ of both input projections and column $j$ of the output projection. For retained indices $I_\ell$, the resulting module is
\begin{equation}
\operatorname{FFN}'_\ell(z)=W_\ell^d[:,I_\ell]
\left[\sigma(W_\ell^g[I_\ell,:]z)\odot(W_\ell^u[I_\ell,:]z)\right].
\end{equation}
The tensors are physically resized, so the model runs with smaller dense operations. We report the cumulative channel-removal ratio
\begin{equation}
\rho=1-\frac{\sum_\ell |I_\ell|}{\sum_\ell m_\ell}.
\label{eq:ratio}
\end{equation}
Because only FFN channels are removed, $\rho$ exceeds the total parameter reduction reported in Table~\ref{tab:efficiency}.

\subsection{Channel Importance Scoring}
\label{app:channel-score-aggregation}
The calibration batch $\mathcal B$ in Eq.~\eqref{eq:channel-taylor-components} is $\mathcal B_{\mathrm{last}}$, the last training batch of anchored OPD, which contains anchor contexts $c^T$ with sampled student responses $\widehat a$. Before each subsequent cut, we recompute the gradient of $\mathcal L_{\mathrm{AOPD}}(\theta;\mathcal B_{\mathrm{last}})$ at the recovered parameters with one forward--backward pass over the batch, which requires no additional generation or environment interaction. The loss averages the reverse KL over all supervised response tokens in the batch, including reasoning and action tokens and excluding prompt and padding tokens. Because recovery trains LoRA adapters, $W$ and $G$ in Eq.~\eqref{eq:channel-taylor-components} refer to the effective FFN weights with the adapters merged; each cut removes the selected channels from both the base weights and the adapters, which continue training in the next stage.

\paragraph{Initial calibration.}
For the initial cut, we add Gaussian noise to the FFN weights of a separate student copy, with standard deviation $0.01$ times each weight matrix's RMS. This perturbed copy generates the anchored calibration responses and supplies the gradients for initial channel scoring. The resulting channel selection is applied to the unperturbed student; later calibration stages use the recovered student without this perturbation.

\subsection{Baseline Adaptations for Uniform Layer-wise Pruning}
\label{app:baseline-adaptations}
All methods, including TAP, use the same target FFN channel-removal ratio in every Transformer layer, so that comparisons isolate channel selection and recovery. For LLM-Pruner \citep{ma2023llmpruner}, we use first-order element-wise importance (Element$^1$) with Summation aggregation, followed by logits KD recovery. For CFSP, we replace its importance-based inter-block sparsity allocation with the uniform target while retaining its within-layer FFN channel-selection criterion \citep{wang2025cfsp}. For SlimGPT, we replace its depth-dependent pruning allocation with the uniform target while retaining its FFN pruning and local reconstruction procedure \citep{ling2024slimgpt}. The baseline names in our tables refer to these adaptations under a shared FFN allocation.

\subsection{Recovery Objectives for Ablation Studies}
\label{app:recovery-objectives}
Let $\pi_T$ be the frozen teacher and $\pi_\theta$ the student. A response $a=(y_1,\ldots,y_{|a|})$ is conditioned on an interaction context $c=(h_{<t},o_t)$, with token prefix $p_j=(c,y_{<j})$ and prompt formatting implicit. For a batch $\mathcal B$ of context--response pairs, let $N_{\mathcal B}=\sum_{(c,a)\in\mathcal B}|a|$ count valid response tokens. All objectives exclude prompt, environment-observation, and padding tokens from supervision.

\paragraph{Logits KD.}
We use fixed teacher-generated pairs $\mathcal B_T=\{(c^T,a^T)\}$ and match the teacher's full-vocabulary distribution on teacher response prefixes $p_j^T=(c^T,y_{<j}^T)$ using forward KL:
\begin{equation}
\mathcal L_{\mathrm{KD}}(\theta)
=\frac{1}{N_{\mathcal B_T}}\sum_{(c^T,a^T)\in\mathcal B_T}\sum_{j=1}^{|a^T|}
D_{\mathrm{KL}}\!\left(\pi_T(\cdot\mid p_j^T)\,\|\,\pi_\theta(\cdot\mid p_j^T)\right).
\label{eq:offline-kd}
\end{equation}

\paragraph{Sequence KD (SeqKD).}
We use the same fixed teacher-generated pairs, but supervise the teacher's response tokens as hard targets:
\begin{equation}
\mathcal L_{\mathrm{SeqKD}}(\theta)
=-\frac{1}{N_{\mathcal B_T}}\sum_{(c^T,a^T)\in\mathcal B_T}\sum_{j=1}^{|a^T|}
\log\pi_\theta(y_j^T\mid p_j^T).
\label{eq:seqkd}
\end{equation}

\paragraph{Vanilla OPD (full-trajectory OPD).}
We sample complete student trajectories $\tau\sim P_{\pi_{\bar\theta},\mathcal E}(\cdot\mid x)$ on training tasks $x$, where $\bar\theta$ denotes the student at sampling time and $\mathcal E$ is the environment. Student actions are executed, so subsequent interaction contexts $c_t^S$ depend on the student's preceding actions and resulting observations. We collect $\mathcal B_S=\{(c_t^S,a_t^S): (c_t^S,a_t^S)\in\tau\}$ from both successful and failed rollouts and minimize reverse KL on student response prefixes $p_{t,j}^S=(c_t^S,y_{t,<j}^S)$:
\begin{equation}
\mathcal L_{\mathrm{OPD}}(\theta)
=\frac{1}{N_{\mathcal B_S}}\sum_{(c_t^S,a_t^S)\in\mathcal B_S}\sum_{j=1}^{|a_t^S|}
D_{\mathrm{KL}}\!\left(\pi_\theta(\cdot\mid p_{t,j}^S)\,\|\,\pi_T(\cdot\mid p_{t,j}^S)\right).
\label{eq:trajectory-opd}
\end{equation}

\subsection{Run Configuration and Data Splits}
\label{app:run-config}
Calibration and recovery use training tasks only. The teacher pool is fixed across methods and stages. Response and episode caps, the action parser, and context handling are shared across comparisons.

\paragraph{TAP recovery and calibration settings.}
Anchored OPD samples 16 prompts per batch and generates one response per prompt with temperature $0.7$ and top-$p=0.9$. LoRA uses rank 8, alpha 16, and dropout 0. We optimize the LoRA parameters with AdamW at a constant learning rate of $10^{-5}$. KL temperature is 1, independently of the generation temperature. Supervision covers valid response tokens and excludes prompt, environment-observation, and padding tokens. For Taylor calibration, we use 16 anchored responses at the first cut and the last batch of 16 recovery responses before each subsequent cut.

\subsection{Recovery Budgets and Controlled Comparisons}
\label{app:recovery-controls}
\label{app:resource-accounting}
Within each model--environment--removal setting, all distillation-based recovery methods in Tables~\ref{tab:main} and~\ref{tab:recovery} share the same supervised budget $Q$ (Table~\ref{tab:token-budgets}), an effective batch size of 16 prompt--response pairs, and the same LoRA, optimizer, and KL-temperature settings as TAP (Appendix~\ref{app:run-config}). The pruning baselines compute channel importance and reconstruction statistics on the same teacher trajectories that form TAP's anchor pool, and full-trajectory OPD samples its rollouts with the same temperature and top-$p$ as anchored OPD. $Q$ is the total number of response tokens obtained by splitting the selected teacher episodes into per-turn prompt--response pairs. It includes reasoning, action, and naturally terminating EOS tokens, and excludes prompts and padding. Every method stops after $Q$ supervised response tokens; for anchored OPD, these are student-response tokens, and the final batch is masked to the remaining budget. Iterative pruning divides $Q$ equally across three stages, each with $20\%$ KD warm-up and $80\%$ anchored OPD.

Table~\ref{tab:recovery} compares recovery recipes under the same cubic schedule and total response-token budget. The OPD recipes use $20\%$ KD warm-up followed by $80\%$ reverse-KL OPD within each stage; the KD control uses the entire stage budget for forward-KL distillation. SeqKD allocates the full stage budget to response-only cross-entropy on fixed teacher-generated trajectories, using hard teacher tokens as targets.

\begin{table}[t]
\centering
\caption{Recovery budget for each main-table setting. All distillation-based methods within a setting share the supervised response-token budget $Q$; total tokens also count the unsupervised prompts.}
\label{tab:token-budgets}
\small
\begin{tabular}{lllrrr}
\toprule
Model & Environment & Removal & Episodes & Response tokens $Q$ (M) & Total tokens (M) \\
\midrule
1.5B & ALFWorld & $40\%$, $60\%$ & 500 & 0.86 & 5.65 \\
1.5B & WebShop & $40\%$ & 500 & 0.18 & 3.10 \\
1.5B & WebShop & $60\%$ & 1,000 & 0.35 & 6.18 \\
\midrule
7B & ALFWorld & $60\%$, $80\%$ & 500 & 0.48 & 3.72 \\
7B & WebShop & $60\%$ & 500 & 0.60 & 3.96 \\
7B & WebShop & $80\%$ & 1,000 & 1.21 & 7.96 \\
\bottomrule
\end{tabular}
\end{table}

\subsection{Efficiency Measurement Protocol}
\label{app:efficiency}
\paragraph{Model inference.}
We benchmark dense and pruned models on the same NVIDIA RTX A6000 GPU using BF16, PyTorch 2.8.0, and Transformers 4.57.3 with SDPA. Each input contains 512 tokens, with batch size 1 and 128 generated tokens. We average three synchronized measurements after one warm-up, excluding model loading. All models receive identical inputs and generate the full output length. Prefill measures prompt processing; decoding averages the subsequent 127 cached generation steps.

\paragraph{Agent execution and cost per success.}
Agent evaluations use vLLM on the same A6000 host with matched evaluator settings and task sets, pooling evaluation time and success counts. End-to-end time $T$ spans initialization, model loading, inference, environment execution, and output collection. Each evaluation uses one GPU. For $N$ attempted tasks and $S$ successes, we report amortized time per task $T/N$ and cost per success $T/S$, the GPU time per successful task including failed attempts. End-to-end speedup is the dense-to-pruned ratio of $T/N$; relative CPS is the pruned-to-dense ratio of $T/S$. These are aggregate costs over each full evaluation.

\end{document}